\PassOptionsToPackage{dvipsnames}{xcolor}
\documentclass{article}

\usepackage{microtype}
\usepackage{graphicx}
\usepackage{subfigure}
\usepackage{booktabs} 

\usepackage{hyperref}

\usepackage[accepted]{icml2024}

\usepackage{amsmath}
\usepackage{amssymb}
\usepackage{mathtools}
\usepackage{amsthm}

\usepackage[capitalize,noabbrev]{cleveref}

\usepackage[font=small,skip=20pt]{caption}

\usepackage{tikz}
\usepackage{tikzscale}
\usepackage{tcolorbox}

\usetikzlibrary{external}
\usetikzlibrary{cd}
\tikzcdset{
  arrow style=tikz,
  diagrams={>={Straight Barb}}
}

\usetikzlibrary{patterns,positioning,arrows,shapes,shapes.geometric}
\usetikzlibrary{decorations,decorations.pathmorphing,intersections,backgrounds}

\usepackage{pgfplots}
\DeclareUnicodeCharacter{2212}{−}
\usepgfplotslibrary{groupplots,dateplot}
\usetikzlibrary{shapes.arrows}
\pgfplotsset{compat=newest}
\usepgfplotslibrary{groupplots}
\pgfplotsset{compat=1.11,
  /pgfplots/ybar legend/.style={
  /pgfplots/legend image code/.code={%
    \draw[##1,/tikz/.cd,yshift=-0.25em]
    (0cm,0cm) rectangle (3pt,0.8em);},
  },
}

\theoremstyle{plain}

\theoremstyle{definition}

\theoremstyle{remark}

\newcommand{\thetadim}{\nu}
\newcommand{\datadim}{n}

\newcommand{\featidx}{i}
\newcommand{\layeridx}{j}

\newcommand{\numhlayers}{\eta}

\newcommand{\featc}{r}
\newcommand{\numfeats}[1]{r_{#1}}
\newcommand{\numnfeats}[1]{\rho_{#1}}

\newcommand{\thetavec}{\boldsymbol\theta}
\newcommand{\muvec}{\boldsymbol\mu}
\newcommand{\fvec}{\mathbf{f}}
\newcommand{\xvec}{\mathbf{x}}
\newcommand{\yvec}{\mathbf{y}}

\newcommand{\Thetamat}{\boldsymbol\Theta}
\newcommand{\Sigmamat}{\boldsymbol\Sigma}
\newcommand{\Psimat}{\boldsymbol\Psi}
\newcommand{\Cmat}{\mathbf{C}}
\newcommand{\Fmat}{\mathbf{F}}
\newcommand{\Gmat}{\mathbf{G}}
\newcommand{\Jmat}{\mathbf{J}}
\newcommand{\Kmat}{\mathbf{K}}
\newcommand{\Xmat}{\mathbf{X}}
\newcommand{\Ymat}{\mathbf{Y}}

\newcommand{\Rsp}[1]{\mathbb{R}^{#1}}

\newcommand{\DKL}{\mathbb{D}_{\mathrm{KL}}}

\newcommand{\care}[1]{{\color{black}#1}}

\icmltitlerunning{\care{Position: Bayesian Deep Learning is Needed in the Age of Large-Scale AI}}

\begin{document}

\twocolumn[
\icmltitle{\care{Position: Bayesian Deep Learning is Needed in the Age of Large-Scale AI}}












\begin{icmlauthorlist}
\icmlauthor{Theodore Papamarkou}{uom}
\icmlauthor{Maria Skoularidou}{broad}
\icmlauthor{Konstantina Palla}{sp}
\icmlauthor{Laurence Aitchison}{bristol}
\icmlauthor{Julyan Arbel}{inria}
\icmlauthor{David Dunson}{duke}
\icmlauthor{Maurizio Filippone}{kaust}
\icmlauthor{Vincent Fortuin}{hai,tum,mcml}
\icmlauthor{Philipp Hennig}{uot}
\icmlauthor{Jos\'{e} Miguel Hern\'{a}ndez-Lobato}{cam}
\icmlauthor{Aliaksandr Hubin}{uoo,nmbu}
\icmlauthor{Alexander Immer}{ETH}
\icmlauthor{Theofanis Karaletsos}{czi}
\icmlauthor{Mohammad Emtiyaz Khan}{RIKEN}
\icmlauthor{Agustinus Kristiadi}{vi}
\icmlauthor{Yingzhen Li}{icl}
\icmlauthor{Stephan Mandt}{uci}
\icmlauthor{Christopher Nemeth}{uol}
\icmlauthor{Michael A. Osborne}{ox}
\icmlauthor{Tim G. J. Rudner}{nyu_cds}
\icmlauthor{David R\"{u}gamer}{mcml,lmu}
\icmlauthor{Yee Whye Teh}{dm,ox2}
\icmlauthor{Max Welling}{uoa}
\icmlauthor{Andrew Gordon Wilson}{nyu}
\icmlauthor{Ruqi Zhang}{uop}
\end{icmlauthorlist}

\icmlaffiliation{uom}{Department of Mathematics, The University of Manchester, Manchester, UK.}
\icmlaffiliation{broad}{Eric and Wendy Schmidt Center, Broad Institute of MIT and Harvard, Cambridge, USA.}
\icmlaffiliation{bristol}{Computational Neuroscience Unit, University of Bristol, Bristol, UK.}
\icmlaffiliation{inria}{Centre Inria de l'Universit\'e Grenoble Alpes, Grenoble, France.}
\icmlaffiliation{duke}{Department of Statistical Science, Duke University, USA.}
\icmlaffiliation{kaust}{Statistics Program, KAUST, Saudi Arabia.}
\icmlaffiliation{hai}{Helmholtz AI, Munich, Germany.}
\icmlaffiliation{tum}{Department of Computer Science, Technical University of Munich, Munich, Germany.}
\icmlaffiliation{mcml}{\care{Munich Center for Machine Learning, Munich, Germany.}}
\icmlaffiliation{uot}{T\"{u}bingen AI Center, University of T\"{u}bingen, T\"{u}bingen, Germany.}
\icmlaffiliation{cam}{Department of Engineering, University of Cambridge, Cambridge, UK.}
\icmlaffiliation{uoo}{Department of Mathematics, University of Oslo, Oslo, Norway.}
\icmlaffiliation{nmbu}{\care{Bioinformatics and Applied Statistics, Norwegian University of Life Sciences, \r{A}s, Norway.}}
\icmlaffiliation{ETH}{\care{Department of Computer Science, ETH Zurich, Switzerland.}}
\icmlaffiliation{czi}{\care{Chan Zuckerberg Initiative, California, USA.}}
\icmlaffiliation{RIKEN}{Center for Advanced Intelligence Project, RIKEN, Tokyo, Japan.}
\icmlaffiliation{vi}{Vector Institute, Toronto, Canada.}
\icmlaffiliation{icl}{Department of Computing, Imperial College London, London, UK.}
\icmlaffiliation{uci}{Department of Computer Science, UC Irvine, Irvine, USA.}
\icmlaffiliation{uol}{Department of Mathematics and Statistics, Lancaster University, Lancaster, UK.}
\icmlaffiliation{ox}{Department of Engineering Science, University of Oxford, Oxford, UK.}
\icmlaffiliation{sp}{Spotify, London, UK.}
\icmlaffiliation{lmu}{Department of Statistics, LMU Munich, Munich, Germany.}
\icmlaffiliation{dm}{DeepMind, London, UK.}
\icmlaffiliation{ox2}{Department of Statistics, University of Oxford, Oxford, UK.}
\icmlaffiliation{uoa}{Informatics Institute, University of Amsterdam, Amsterdam, Netherlands.}
\icmlaffiliation{nyu_cds}{Center for Data Science, New York University, New York, USA.}
\icmlaffiliation{nyu}{Courant Institute of Mathematical Sciences and Center for Data Science, Computer Science Department, New York University, New York, USA.}
\icmlaffiliation{uop}{Department of Computer Science, Purdue University, West Lafayette, USA}

\icmlcorrespondingauthor{Theodore Papamarkou}{theo.papamarkou@manchester.ac.uk}
\icmlcorrespondingauthor{Maria Skoularidou}{mskoular@broadinstitute.org}
\icmlcorrespondingauthor{Konstantina Palla}{konstantinap@spotify.com}

\icmlkeywords{Bayesian deep learning, ensemble methods, Gaussian process, Laplace approximation, Markov chain Monte Carlo, 
reliable AI, uncertainty quantification}

\vskip 0.3in
]



\printAffiliationsAndNotice{}  

\begin{abstract}
In the current landscape of deep learning research, there is a predominant emphasis on achieving high predictive accuracy in supervised tasks involving large image and language datasets. However, a broader perspective reveals a multitude of overlooked metrics, tasks, and data types, such as uncertainty, active and continual learning, and scientific data, that demand attention. Bayesian deep learning (BDL) constitutes a promising avenue, offering advantages across these diverse settings. This paper posits that BDL can elevate the capabilities of deep learning. It revisits the strengths of BDL, acknowledges existing challenges, and highlights some exciting research avenues aimed at addressing these obstacles. Looking ahead, the discussion focuses on possible ways to combine large-scale foundation models with BDL to unlock their full potential.
\end{abstract}


\section{Introduction}

The roots of Bayesian inference can be traced back to the eighteenth century, with the foundational work of Thomas Bayes in the field of probability theory. Bayes' theorem, formulated posthumously in the 1760s~\citep{bayes1763essay}, laid the groundwork for a probabilistic approach to statistical reasoning. \care{At a high level, Bayes' theorem describes how to update a belief given some evidence.
Formally, Bayes' theorem states the posterior probability density function $p(\thetavec \vert \mathcal{D})$ evaluated at a parameter value $\thetavec\in\Rsp{\thetadim}$ given some evidence (training dataset) $\mathcal{D}$ as a function of three probability density functions, namely the prior $p(\thetavec)$ of $\thetavec$ before evidence $\mathcal{D}$ is considered, the likelihood $p(\mathcal{D} \vert \thetavec)$ of evidence $\mathcal{D}$ given the parameter value $\thetavec$, and the marginal probability density function of evidence $\mathcal{D}$ under any parameter value:}
\begin{equation*}
\care{p(\thetavec \vert \mathcal{D}) =
\frac{p(\mathcal{D} \vert \thetavec) \, p(\thetavec)}{p(\mathcal{D})}.}
\end{equation*}
Over the centuries, Bayesian methods have made a profound impact across various scientific disciplines, offering a principled framework for updating beliefs based on new evidence and accommodating uncertainty in model parameters. From Bayesian statistics in the early twentieth century to the Bayesian revolution in its second half~\citep{Jaynes2003Probability}, the approach has evolved, influencing fields ranging from physics to medicine and artificial intelligence (AI).

\care{The Bayesian view finds many uses in deep learning, including problems of interpretability and characterization of predictive uncertainty. Applications of Bayes' theorem estimate the posterior distribution of neural network (NN) parameters, thus providing a probabilistic understanding and interpretation of the parameters. Furthermore, Bayes' theorem underpins posterior predictive distribution estimation, making it possible to quantify the uncertainty of NN predictions. Interpreting the role of NN parameters and quantifying uncertainty in predictions facilitates risk assessment and improves safety in decision-making.}

\begin{figure}
  \centering
  \resizebox{\linewidth}{!}{\includegraphics{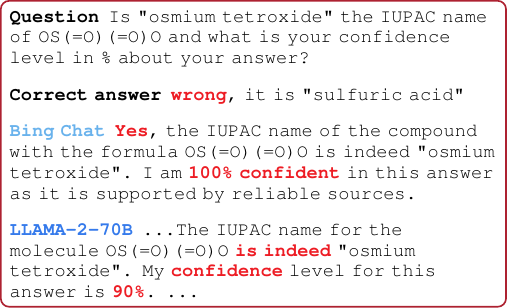}}
  \caption{Popular LLM chat assistants, such as Bing Chat (using GPT-4) and LLAMA-2-70B, often produce \emph{wrong answer} with \emph{very high confidence}, indicating that their confidence is not calibrated. BDL has traditionally been used to overcome this kind of overconfidence problem and yet BDL is underutilized in the LLM era. Note that OS(=O)(=O)O is a textual representation of the well-known molecule H\(_2\)SO\(_4\) and can easily be looked up on Wikipedia. Emphasis and ellipsis ours. Accessed on 2024-01-23.}
  \label{fig:chatbots}
  \vspace{-1.2em}
\end{figure}

In the last two decades, the Bayesian deep learning (BDL) framework, which combines Bayesian principles with deep learning, has garnered significant attention. Despite its potential to provide uncertainty estimates and improve model \care{interpretability}, generalization, and robustness, mainstream adoption of BDL has been sluggish on both the research and application fronts. A primary concern that is often voiced is the lack of scalability of BDL. However, in an era marked by the widespread and rapid adoption of extensively parameterized deep learning models, this paper posits that BDL has untapped potential and can significantly contribute to the current AI landscape. Recognizing the need to revisit the applicability of BDL, especially in the context of largely parameterized deep learning models, this paper aims to critically analyze the existing challenges that hinder the broader acceptance of BDL. By delving into these challenges and proposing avenues for future research, the paper seeks to unlock the full potential of \care{BDL}.

The reason Bayesian concepts are not mainstream in deep learning is not that deep learning makes uncertainty obsolete. In fact, reliable epistemic uncertainty is more relevant than ever in a world of massively overparameterized models. 
For example, out-of-distribution prompts demonstrate that large language models (LLMs) urgently need reliable uncertainty quantification (UQ); see \cref{fig:chatbots}. The problem is that exact Bayesian inference is typically too computationally expensive. 

\textbf{Position.}
\textbf{This position paper argues that the advancement of BDL can overcome many of the challenges that deep learning faces nowadays. Notably, BDL methods can prove instrumental in meeting the needs of the 21st century for more mature AI systems and safety-critical decision-making algorithms that can reliably assess uncertainties and incorporate existing knowledge.} For example, BDL methods can mitigate risks arising from overly confident yet incorrect predictions made by LLMs (see Figure~\ref{fig:chatbots}). The major impediment to the development of broadly adoptable BDL methods is scalability, yet this paper proposes research directions that promise to make BDL more amenable to contemporary deep learning.

\care{Bayesian approaches to deep learning provide several advantages over frequentist alternatives. First, BDL reduces the importance of hyper-parameter tuning through incorporating relevant hyper-priors~\citep{lampinen2001}. Second, in contrast to post-hoc regularization techniques for training on small datasets, BDL enables the use of domain knowledge priors~\citep{sam2024}. Third, BDL approaches to decision-making are more advantageous than frequentist approaches in terms of mitigating the asymmetric costs of errors~\citep{tump2022}. Although there exist non-Bayesian approaches promoting the concept of decision calibration in classification problems, which deal with such asymmetric errors and are suitable for decision-making applications~\citep{zhao2021}, BDL has the added advantage of providing uncertainties over predictions, which can enrich decision-making, for example, by deferring a decision to a later stage when more data is gathered and uncertainty is lower. Fourth, in contrast to conformal prediction, BDL does not require the exchangeability assumption and enables dependence between data across spatiotemporal dimensions through appropriate latent variables~\citep{tran2020}.}

\textbf{Paper structure.}
Section~\ref{sec:why-BDL-matters} explains why BDL matters by highlighting the strengths of BDL. Section~\ref{sec:existing-BDL} critically reflects on the challenges that current BDL methods face. Section~\ref{sec:new-BDL} identifies research directions for the development of scalable BDL methods that can overcome these challenges and become as computationally efficient as established deep learning solutions. The paper concludes with final remarks on the future of BDL (Section~\ref{sec:final-remarks}).
\care{\cref{sec:background} is a self-contained introductory tutorial on the basics of Bayesian methodology and BDL, providing background knowledge on several Bayesian methods discussed in this paper.}

\section{Why Bayesian Deep Learning Matters}
\label{sec:why-BDL-matters}

BDL is a computational framework that combines Bayesian inference principles with deep learning models. Unlike traditional deep learning methods that often provide point estimates, BDL provides a full probability distribution over the parameters, allowing for a principled handling of uncertainty. This intrinsic \textit{uncertainty quantification} is particularly valuable in real-world scenarios where data are limited or noisy. Moreover, BDL accommodates the incorporation of prior information, encapsulated in the choice of a prior distribution. This \textit{integration of prior beliefs} serves as an inductive bias, enabling the model to leverage existing knowledge and providing a principled way to incorporate domain expertise. Based on Bayesian principles, BDL allows \textit{updating beliefs} about uncertain parameters \textit{in light of new evidence}, combining prior knowledge with observed data through Bayes' theorem~\citep{bayes1763essay}. Several works aim to improve the understanding of BDL~\citep{wilson2020bayesian, izmailov2021, izmailov2021dangers, kristiadi2022, papamarkou2022, kapoor2022uncertainty, khan2023, papamarkou2023,rudner2023fsmap}.

BDL has shown substantial potential in a range of critical application domains, such as healthcare~\citep{peng2019bayesian, abdar2021uncertainty, abdullah2022review,lechuga2023m2d2,band2021benchmarking}, single-cell biology~\citep{way2018bayesian}, drug discovery~\citep{gruver2021effective,stanton2022accelerating,gruver2023protein,klarner2023qsavi}, agriculture~\citep{hernandez2020uncertainty}, astrophysics~\citep{soboczenski2018bayesian, ferreira2020galaxy}, nanotechnology~\citep{leitherer2021robust}, physics~\citep{cranmer2021bayesian}, climate science~\citep{vandal2018quantifying, luo2022bayesian}, smart electricity grids~\citep{yang2019bayesian}, wearables~\citep{manogaran2019wearable, zhou2020human}, robotics~\citep{shi2021bayesian, mur2023bayesian}, and autonomous driving~\citep{mcallister2017concrete}. This section outlines the strengths of BDL to motivate the advancement of BDL in the era of large-scale AI.

\subsection{Uncertainty Quantification}

UQ in BDL improves the reliability of the decision-making process and is valuable when the model encounters ambiguous or out-of-distribution inputs~\citep{tran2022plex}. In such instances, the model can signal its lack of confidence in the predictions through the associated probability instead of providing underperforming point estimates. The importance of predictive UQ is especially emphasized in the context of AI-informed decision-making, such as in healthcare~\citep{band2021benchmarking,lechuga2023m2d2}. In safety-critical domains, \textit{reliable UQ} can be used to deploy models more safely by deferring to a human expert whenever an AI system has high uncertainty about its prediction~\citep{tran2022plex,rudner2022fsvi,rudner2023fseb}. This capability is also pertinent to address current challenges in language models, where uncertainty quantification can be used to mitigate risks associated with overly confident but incorrect model predictions~\citep{kadavath2022language}; see~\cref{fig:chatbots} for an example. Similarly, BDL can be useful for modern challenges, such as hallucinations~\citep{ji2023survey} and adversarial attacks~\citep{andriushchenko2023adversarial} in LLMs, or jailbreaking in text-to-image models~\citep{yang2023sneakyprompt}. 

In scientific domains, including but not limited to chemistry and material sciences, where experimental data collection is resource-intensive or constrained, parameter spaces are high-dimensional, and models are inherently complex, BDL excels by providing robust estimates of uncertainty. This attribute is particularly crucial for guiding decisions in inverse design problems, optimizing resource utilization through Bayesian experimental design, optimization, and model selection~\citep{stanton2022accelerating,gruver2023protein,li2023study, Rainforth2023ModernBE, bamler2020augmenting,lotfi2022bayesian,immer2021scalable,immer2023stochastic}.

\subsection{Data Efficiency}

\care{BDL has manifested data efficiency in various contexts. Notably, BDL methods have been developed for few-shot learning scenarios~\citep{yoon2018,patacchiola2020} and for federated learning under limited data~\citep{zhang2022}.}

Unlike many machine learning approaches that may require large datasets to generalize effectively, BDL leverages prior knowledge and updates beliefs as new data become available. This allows BDL to extract meaningful information from \textit{small datasets}, making it more efficient in scenarios where collecting large amounts of data is challenging or costly~\citep{finzi2021residual,immer2022invariance,shwartz2022,schwobel2022last,van2023learning}. In addition, the \textit{regularization} effect introduced by the probabilistic nature of its Bayesian approach is beneficial in \textit{preventing overfitting} and contributing to better \textit{generalization} from fewer samples~\citep{rothfuss2022pac, sharma2023incorporating}. BDL's uncertainty modeling helps resist the influence of outliers, making it well-suited for real-world scenarios with noisy or out-of-distribution data. This also makes it attractive for foundation model fine-tuning, where data are commonly small and sparse, and uncertainty is important.

Furthermore, the UQ capabilities of BDL allow for an informed selection of data points for labeling. Using prior knowledge and continually updating beliefs as new information arrives, BDL optimizes the iterative process of \textit{active learning}, strategically choosing the most informative instances for labeling to enhance model performance~\citep{galAL2017}. This capability may be particularly advantageous for addressing the current challenge of efficiently selecting demonstrations in in-context learning scenarios~\citep{margatina-etal-2023-active} or fine-tuning with human feedback~\citep{casper2023open}.

\subsection{Adaptability to New and Evolving Domains}
\label{subsec:adaptability}

By dynamically updating prior beliefs in response to new evidence, BDL allows selective retention of valuable information from previous tasks while adapting to new ones, thus improving \textit{knowledge transfer} across diverse domains and tasks~\citep{rothfuss2021pacoh,rothfuss2022pac,rudner2024uap}. This is crucial for developing AI systems that can adapt to new situations or temporally evolving domains~\citep{nguyen2018variational,rudner2022sfsvi}, as in the case of continual or lifelong learning. The contrast with traditional approaches in large-scale machine learning becomes apparent, as these static models assume that the underlying patterns in the data remain constant over time and struggle with the constant influx of new data and changes in underlying patterns.
 
\subsection{Model Misspecification and Interpretability}

Bayesian model averaging (BMA) acknowledges and quantifies uncertainty in the choice of model structure. Instead of relying on a single fixed model, BMA considers a distribution of possible models~\citep{hoeting1998bayesian,hoeting1999bayesian, wasserman2000bayesian}. By incorporating model priors and inferring model posteriors, BDL allows BMA to calibrate uncertainty over network architectures~\citep{hubin2019combining,skaaret2023sparsifying}. By averaging predictions over different model possibilities, BMA \textit{attenuates} the impact of \textit{model misspecification}, offering a robust framework that accounts for uncertainty in both parameter values and model structures, ultimately leading to more reliable and interpretable predictions~\citep{hubin2021flexible, wang2023m2ib,bouchiat2023laplace}.

The interpretation of parameters and structures may seem less crucial in BDL, where overparameterized neural networks serve as functional approximations to unknown data-generating processes. However, research is required to establish reproducible and interpretable Bayesian inferences from deep neural networks (DNNs), especially in applications where black-box prediction is not the primary objective, particularly in scientific contexts~\citep{rugamer2023,wang2023m2ib,dold2024}. BMA-centric research in BDL can be valuable in these directions.

\section{Current Challenges}\label{sec:existing-BDL}

\begin{figure}
\centering
\includegraphics[width=\linewidth]{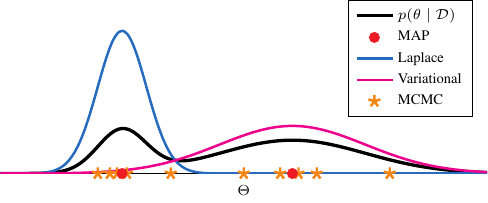}
\caption{Different BDL methods for approximating a posterior $p(\theta \mid \mathcal{D})$ on a parameter space $\Theta$.
While Laplace and Gaussian-based variational approaches yield Gaussian approximations, they generally capture different local modes of the posterior. Ensemble methods use maximum a posteriori estimates as their samples.}
\label{fig:bld_methods}
\vspace{-1em}
\end{figure}

One of the challenges in BDL is the computational cost incurred~\citep{izmailov2021}. Despite the BDL advantages outlined in Section~\ref{sec:why-BDL-matters}, within the realm of Bayesian approaches, Gaussian Processes (GPs) remain the preferred choice in computationally demanding scenarios such as scientific discovery~\citep{tom2023calibration,griffiths2023gauche,strieth2023laser}. Showing that BDL works cheaply, or at least with practical efficiency under modern settings in the real world, is one of the most important problems that remains to be addressed.  
This section aims to explore the complexities of BDL, highlighting two main challenges that contribute to its difficulties in deployment: posterior computation (\cref{fig:bld_methods}) and prior specification. It is also explored how scalability arises as a main challenge in BDL. The section concludes with difficulties in the adoption of BDL in foundation models. Challenges related to the lack of convergence and performance metrics and benchmarks for BDL are discussed in~\cref{sec:diagnostics}.

\subsection{Laplace and Variational Approximations}

Laplace and variational approximations use geometric or differential information about the empirical loss to construct closed-form (usually Gaussian) probability measures to approximate the posterior. Despite their simple nature and long history~\citep{mackay1992bayesian}, they often show competitive predictive performance~\citep{daxberger2021a,rudner2022fsvi,antoran2023,rudner2023fseb}. More importantly, their closed-form nature, leveraging automatically computed differential quantities and the foundations of numerical linear algebra, allows theoretical analysis~\citep{pmlr-v119-kristiadi20a} and analytical functionality, such as calibration~\citep{pmlr-v161-kristiadi21a,kristiadi2021infinite} and marginalization~\citep{khan2019approximate,immer2021scalable,pmlr-v130-immer21a}, which are less elegant with stochastic approaches. Laplace-approximated neural networks~\citep{ritter2018} are particularly tempting because they add no computational cost during training, and require limited computational overhead (comparable to a few epochs) for post-hoc UQ. 
Moreover, recent variational objectives~\citep{alemi2023} provide alternative means of prediction that avoid internal marginalization. 

Alternatively, SWAG~\citep{maddox2019simple} is another scalable approximation that creates a Gaussian approximate posterior from stochastic gradient descent (SGD) iterations~\citep{mandt2017stochastic} with a modified learning rate schedule. Similarly to the Laplace approximation, it does not cost much more than standard training. However, SWAG estimates curvature from the trajectory of SGD, rather than the Hessian at a single point. By producing a deterministic probability measure from stochastic gradients, it bridges the gap between deterministic and stochastic procedures.

Despite their analytic strengths, these approximations remain fundamentally local, capturing only a single mode of the multimodal Bayesian neural network (BNN) posterior. Arguably, their most fundamental flaw is that their posterior is dependent on the parametrization of the BNN~\citep{mackay1998choice} and thus inconsistent with some of the most basic properties of probability measures~\citep{kristiadi2023the}.
Furthermore, the local posterior geometry may be poorly approximated by a Gaussian distribution, which can lead to underconfidence when sampling from the Laplace approximation~\citep{lawrence2001variational}, a problem that can be mitigated by linearization~\citep{pmlr-v130-immer21a}.

\subsection{Ensembles}

Deep ensembling involves the retraining of an NN with various initializations, followed by averaging the resulting models. It is effective in approximating the posterior predictive distribution~\citep{wilson2020bayesian}. Recent theoretical advances have established precise connections between ensembles and Bayesian methods~\citep{ciosek2020conservative, he2020bayesian,wild2023a}.

An open question in BDL is whether one can develop scalable Bayesian inference methods that outperform deep ensembles.~\citet{izmailov2021} have shown that Hamiltonian Monte Carlo (HMC) often outperforms deep ensembles, but with significant additional computational overhead.

When dealing with large and computationally expensive deep learning models, such as LLMs, the use of deep ensembles may encounter significant challenges due to the associated training and execution costs. Therefore, these large models may motivate research into more efficient architectures and inference paradigms, such as posterior distillation or repulsive ensembles~\citep{dangelo2021repulsive}, to improve uncertainty calibration and sparser model use. 

\subsection{Posterior Sampling Algorithms}

Within the realm of Markov chain Monte Carlo~\citep[MCMC;][]{brooks2011handbook} for BDL, stochastic gradient MCMC~\citep[SG-MCMC;][]{nemeth2021stochastic} algorithms, such as stochastic gradient Langevin dynamics~\citep[SG-LD;][]{welling2011bayesian} and stochastic gradient HMC~\citep[SG-HMC;][]{chen2014stochastic}, have emerged as widely adopted tools. Despite offering improved posterior approximations, SG-MCMC algorithms exhibit slower convergence compared to SGD~\citep{Robbins1951ASA}. This deceleration results from the increased iterations required by SG-MCMC to thoroughly explore the posterior distribution beyond locating the mode. 

Furthermore, SG-MCMC is still considered expensive for deep learning applications. A step forward in this regard would be to learn from the machine learning and systems community how to make Monte Carlo faster using contemporary hardware~\citep{zhang2022low,wang2023enhancing}. 
Algorithms such as Stein variational gradient descent~\citep[SVGD;][]{liu2016stein} occupy a middle ground between optimization and sampling, by employing optimization-type updates but with a set of interacting particles. While recent advances show promising results in BNN settings~\citep{dangelo2021stein, dangelo2021repulsive, pielok2022approximate}, these methods often perform poorly in high-dimensional problems.
Alternatively, convergence rates and posterior exploration can be improved with cyclical step-size schedules~\citep{zhang2019cyclical}.

However, despite these advances, the persistent challenges posed by the highly multimodal and high-dimensional nature of BDL posteriors continue to impede the accurate characterization of the full posterior distribution via sampling. There is a need for SG-MCMC algorithms that not only match the speed of SGD, as deployed for optimization in typical deep learning settings, but also deliver high-quality approximations of the posterior to ensure practical utility. 

\subsection{Prior Specification}

The prior over parameters induces a prior over functions, and it is the prior over functions that matters for generalization~\citep{wilson2020bayesian}. Fortunately, the structure in neural network architectures already endows this prior over functions with many desirable properties, such as translation equivariance if a CNN architecture is used. At the same time, defining priors over the parameters is hindered by the complexity and unintelligibility of high-dimensional spaces in BDL. Thus, one aim is to construct informative proper priors on neural network weights that are computationally efficient and favor solutions with desirable model properties~\citep{vladimirova2019bayesian,vladimirova2021accurate,fortuin2022bayesian,rudner2023fseb}, such as priors that favor models with reliable uncertainty estimates~\citep{rudner2024uap}, a high degree of fairness~\citep{rudner2024gap}, generalization under covariate shifts~\citep{klarner2023qsavi}, equivariance~\citep{finzi2021residual}, or a high level of sparsity~\citep{ghosh2018structured,polson2018posterior,hubin2019combining}.
Weight priors can be cast as neural fields using low-dimensional unit latent variables~\citep{karaletsos2018probabilistic, karaletsos2020hierarchical} paired with hypernetworks or GPs to express prior knowledge about the field, thus omitting direct parameterizations of beliefs over weights in favor of geometric or other properties of units. 

Recent research has developed priors in function space rather than in weight space~\citep{tran2022,rudner2022sfsvi,rudner2023fsmap}. Function-space priors also raise some issues, such as ill-defined variational objectives~\citep{burt2020understanding,rudner2022fsvi} or, in some cases, the need to perform computationally costly GP approximations. There are alternative ways to specify function-space priors beyond GPs. For example, informative function-space priors may be constructed through self-supervising learning~\citep{shwartz2022, sharma2023incorporating}. 

\subsection{Scalability}
\label{sec:scalability}

The presence of symmetries in the parameter space of NNs yields computational redundancies~\citep{wiese2023}. Addressing the complexity and identifiability issues arising from these symmetries in the context of BDL can significantly impact scalability. Proposed solutions involve the incorporation of symmetry-based constraints in BDL inference methods~\citep{sen2024} or the design of symmetry-aware priors~\citep{atzeni2023infusing}. 
However, removing symmetries may not be an optimal strategy, since part of the success of deep learning can be attributed to the overparameterization of NNs, allowing rapid exploration of numerous hypotheses during training or having other positive `side effects' such as induced sparsity~\citep{kolb2023}. 

Contrary to the misconception that BNNs inherently suffer from limitations in speed and memory efficiency compared to deterministic NNs, recent advances challenge this notion. \care{For instance}, research by~\citet{ritter2021} shows that BNNs can achieve up to four times greater memory efficiency than their deterministic counterparts in terms of the number of parameters. Furthermore, strategies such as recycling the standard training trajectory to construct approximate posteriors, as proposed by~\citet{maddox2019simple}, incur negligible additional computation costs. Hybrid models that combine NNs with GPs, such as deep kernel learning~\citep[DKL;][]{wilson2016deep}, are also \care{only} marginally slower or more memory-consuming than deterministic NNs. 

Although UQ is important across various domains, it should not come at the cost of reduced predictive performance. BDL must strike a balance by ensuring that the computational cost of UQ matches that of point estimation. Otherwise, investing computational resources to improve the predictive performance of deep learning models might be a more prudent option. Some may contend that ensembles are less affected by this concern due to their embarrassingly parallel nature. However, in an era where even industry leaders encounter limitations in graphics processing unit (GPU) resources required to train a single large deep learning model, relying solely on parallelism becomes inadequate. 
Simultaneously achieving time efficiency, memory efficiency, and high model utility (in terms of predictive performance and uncertainty calibration) remains the grand challenge; this is the holy grail of approximate Bayesian inference. 

\subsection{Foundation Models}

Deep learning is in the midst of a paradigm shift into the `foundation model' era, characterized by models with billions, rather than millions, of parameters, with a predominant focus on language rather than vision. BDL approaches to LLMs are relatively unexplored, both in terms of methods and applications. While state-of-the-art approximate inference algorithms can effectively handle models with millions of parameters, only a limited number of works have considered Bayesian approaches to LLMs~\citep{xie2021,cohen2022bayesian, margatina2022}. 
\care{In particular, some BDL methods for LLMs have been developed by using Bayesian low-rank adaptation~\citep[LoRA;][]{yang2023bayesian, onal2024gaussian}, Bayesian optimization~\citep{kristiadi2024sober}, and Bayesian reward modeling~\citep{yang2024reward}.}

As discussed in~\cref{sec:why-BDL-matters}, BDL emerges as a solution to address limitations in foundation models, particularly in scenarios where data availability is limited. In contexts involving personalized data~\citep{moor2023foundation} or causal inference applications~\citep{zhang2023towards}, such as individual treatment effect estimation, where small datasets prevail, the capacity of BDL for uncertainty estimation aligns seamlessly. The fine-tuning settings of foundation models in small-data scenarios is another example. 
\care{While foundation models are few-shot learners~\citep{brown2020}, BDL offers interpretable uncertainty quantification, which is particularly important in data-limited settings. Moreover, BDL facilitates predictive uncertainty estimation and robust decision-making under uncertainty.}

Foundation models represent a valuable frontier for BDL research, particularly around evaluation and applications. What applications of LLMs or transformers are going to benefit from Bayesian inference tools, such as marginalization and priors? More generally, more meaningful applications are needed to convincingly demonstrate that BDL principles go beyond proof-of-concept.
The representation of epistemic uncertainty will possibly be most valuable when LLMs or other large-scale NNs are deployed in settings outside of the realm of their training data. For example, Bayesian approaches can be developed and tested in the time series context of applying LLMs in downstream forecasting tasks~\citep{gruver2023}. 

\section{Proposed Future Directions}\label{sec:new-BDL}

This section, driven by the challenges described in~\cref{sec:existing-BDL}, presents ongoing research initiatives dedicated to addressing these challenges, particularly focusing on scalability. Subsection~\ref{subsec:other_future_directions} presents more recent or less widely studied Bayesian research approaches to deep learning. Some topical developments in BDL are discussed in~\cref{sec:topical}.

\subsection{Posterior Sampling Algorithms}

There is a need for new classes of posterior sampling algorithms that perform better on deep neural networks (DNNs). These algorithms should aim to enhance efficiency, reduce computational overhead, and enable more effective exploration of high-dimensional parameter spaces.

SG-MCMC with tempered posteriors may potentially overcome the issue of sampling from multiple modes. This could be achieved by developing new sampling approaches that can be based on ideas from optimal transport theory~\citep{villani2021topics}, score-based diffusion models~\citep{song2020score}, and ordinary differential equation (ODE) approaches such as flow matching~\citep{lipman2022flow}, which use NNs to learn a mapping from a simpler (usually Gaussian) distribution to a complex data distribution (for example, a distribution of images). So, one could plausibly use an NN either to learn a mapping between the BDL posterior and a Gaussian distribution or to use an NN in an MCMC proposal mechanism. 

Generally, instead of just focusing on local information about the posterior, there is a need for SG-MCMC algorithms that are able to move rapidly across isolated modes, for instance, using normalizing flows. Since one may not expect to accurately approximate a high-dimensional posterior with respect to all the BNN parameters, novel performance metrics may target lower-dimensional functionals of interest, including UQ as a key piece.

One approach is to incorporate appropriate constraints to attain identifiability, for instance, by making inference on the latent BNN structure~\citep{gu2023}. Instead, one can focus on identifiable functionals for canonical classes of NNs, targeting posterior approximation algorithms for these functionals. Further, one may consider decoupling approaches, which use the BNN as a black box to fit the data-generating model and then choose appropriate loss functions to conduct inference in a second stage. 

Another promising approach is running SG-MCMC algorithms in subspaces of the parameter space, for example, linear or sparse subspaces~\citep{izmailov2020,li2023training}, further enabling the formulation of uncertainty statements for targeted subnetworks~\citep{dold2024}. In the future, SG-MCMC operating on QLoRA~\citep{dettmers2023} or non-linear subspaces may be constructed. Besides treating subspaces deterministically, posterior dependencies between subspaces can be broken systematically, leading to novel hybrid samplers that combine structured variational inference with MCMC~\citep{alexos2022structured} to achieve compute-accuracy trade-offs.
Subsampling for BDL can be combined with reasoning about transfer learning~\citep{kirichenko2023last}. 

\subsection{Hybrid Bayesian Approaches}

In the future, practical BDL approaches may capture uncertainty over a limited part of the model, while other parts may be estimated efficiently using point estimation. So, one may consider hybrid approaches that combine Bayesian methods with the efficiency of deterministic deep learning.

This could involve developing methods that selectively apply Bayesian approaches in critical areas of the model where capturing uncertainty will be more useful and cheaper, while maintaining a deterministic approach for other parts of the model~\citep{daxberger2021a}. The last-layer Laplace approximation is an example of this~\citep{daxberger2021b}. Such hybrid approaches are a promising area for future research.

Combinations of deep learning methods and GPs have traditionally been limited by the lack of scalability of GPs. However, recent advances in scaling up GP inference are promising for making these hybrid models more widespread. 
DKL~\citep{wilson2016deep} is one example of such a hybrid model. The DKL scalability frontier may be further pushed by exploiting advances in GP scalability.

There exists a prolific literature on connecting BDL and deep Gaussian processes~\citep[DGPs;][]{wilson2012gaussian,damianou2013,agrawal2020}. This line of work involves neural network GPs~\citep{neal1996,matthews2018}, which are GPs that arise as infinite-width limits of NNs. Theoretical insights into BDL may come from the connection between NNs and GPs. 

\subsection{Deep Kernel Processes and Machines}

Deep kernel processes (DKPs) constitute a family of deep non-parametric approaches to BDL~\citep{aitchison2021deep,ober2021variational,ober2023improved}. A DKP is a DGP, in which one treats the kernels, rather than the features, as random variables. It is possible to derive the prior and perform inference for kernels, without needing DGP features or BNN weights~\citep{aitchison2021deep}. Thus, DKPs avoid the highly multimodal posteriors caused by permutation symmetries in BDL. It is challenging to accurately approximate these multimodal posteriors with simplified parametric families, for instance, as used in Laplace or variational inference. In contrast, the DKP posterior in practice tends to be unimodal~\citep{yang2023theory}. DKPs are a generalization of kernel inverse Wishart processes~\citep{shah2014student}, but with non-linear transformations of the kernel, which are useful in representation learning.

Deep kernel machines~\citep[DKMs;][]{milsom2023convolutional,yang2023theory} go further, by taking the infinite-width limit of a DKP. Usually such an infinite-width limit would eliminate representation learning. However, DKMs carefully temper the likelihood in order to retain representation learning, and are thereby able to attain state-of-the-art predictive performance~\citep{milsom2023convolutional}, while their theoretical implications are profound for BDL. DKMs offer key insights into what `inference in function space' really means and how it relates to representation learning. Specifically, the kernels learned at every layer in a DKM define a `function space' at every layer. In fact, in a DKM, the true posterior over features is multivariate Gaussian with covariance given by the learned kernel~\citep{aitchison2021deep}. Representation learning occurs as these function spaces at every layer are modulated by training to focus on the features that matter for predictive performance.

\subsection{Semi-Supervised and Self-Supervised Learning}

From a Bayesian perspective, one of the surprises in modern deep learning has been the success of semi-supervised learning, where the objective is seemingly arbitrary (or at least, it does not obviously correspond to a likelihood in a known model). Additionally, in Bayesian inference, there are phenomena such as the `cold posterior effect'~\citep{aitchison2020statistical,wenzel2020}, in which BDL appears to attain more competitive predictive performance by taking the posterior to a power greater than one, thereby shrinking the posterior. In particular, the patterns exploited by semi-supervised learning arise from data curation~\citep{ganev2023}. If semi-supervised learning is performed on uncurated data, any improvements disappear. This casts doubt on the applicability of semi-supervised learning on real-world uncurated datasets. The cold posterior results can also be explained by underconfident aleatoric uncertainty representation~\citep{kapoor2022uncertainty}.  

Self-supervised learning is an alternative to semi-supervised learning. Self-supervised learning is based on objectives such as mutual information between latent representations of two augmentations of the same underlying image. From a Bayesian perspective, these objectives appear to be ad hoc, as they do not correspond to any likelihood. However, it is possible to formulate a rigorous likelihood in the form of a recognition-parameterized model~\citep{aitchison2023}. This provides insight into the workings of self-supervised learning and how to generalize it to new settings, such as viewing it as a way to learn Bayesian priors~\citep{shwartz2022,sharma2023incorporating}. 

\subsection{Mixed Precision and Tensor Computations}

The success of deep learning is closely tied to its coupling with modern computing and specialized hardware, leveraging technologies like GPUs. Recent investigations within deep learning on the impact of mixed precision point to a role for Bayes, particularly probabilistic numerics~\citep{oates2019modern}, in making more efficient use of computation. Mixed precision introduces uncertainty into the internal computations of a model, which Bayes can effectively propagate to downstream predictions. Furthermore, mixed precision requires making decisions about which precision to use, where Bayes can ensure that these decisions are optimal and sensitive to the relations between numerical tasks. 
Drawing inspiration from specialized hardware, such as tensor processing units, there is potential for a similar trajectory in BDL to address scalability concerns~\citep{mansinghka2009natively}. This suggests that the creation of dedicated hardware for BDL has the potential to spark a reevaluation of inference strategies. 

In a parallel vein, accelerating software development is crucial to encouraging deep learning practitioners to adopt Bayesian methods. There is a demand for user-friendly software that facilitates the integration of BDL into various projects. The goal is to make BDL usage competitive in terms of human effort compared to standard deep learning practices. For details on BDL software efforts, see~\cref{sec:software}.

\subsection{Compression Strategies}

To decrease the computational cost of BDL models, for both memory efficiency and computational speed, compression strategies are being explored. An approach involves using sparsity-inducing priors to prune large parts of BNNs~\citep{louizos2017bayesian}. Alternatively, the prior can serve as an entropy model, enabling the compression of BNN weights~\citep{yang2023introduction}. Methods such as relative entropy coding and variational Bayesian quantization, where the quantization grid is dynamically refined, provide efficient BNN compression~\citep{yang2020}. These novel tools could also be used to dynamically decode a Bayesian ensemble at test time to various levels of precision or ensemble size, resulting in precision-compute trade-offs. 

Furthermore, in the context of compressing NN weights, a viable approach involves obtaining the posterior distribution based on observed data and encoding a sample into a bit sequence to send to a receiver~\citep{havasi2018}.  The receiver can then extract the posterior sample and use the corresponding weights to make predictions. In practice, approximations are needed to obtain the posterior, encode the sample, and use the corresponding weights to make predictions. Despite the need for approximations in the process, this method yields commendable trade-offs between compression cost and predictive quality compared to alternatives centered on deterministic weight compression.

\subsection{Other Future Directions}
\label{subsec:other_future_directions}

\textbf{Bayesian transfer and continual learning.}
The transfer learning paradigm is quickly becoming a standard way to deploy deep learning models. As noted in \care{Subsection~\ref{subsec:adaptability}}, BDL is optimized for transfer learning. The focus is not solely on transferring an initialization as in traditional deep learning; instead, knowledge of the source task may inform the shapes and locations of optima on downstream tasks~\citep{shwartz2022, rudner2022sfsvi,rudner2023fseb}. Self-supervised learning can also be used to create informative self-supervised priors for transfer learning~\citep{shwartz2022, sharma2023incorporating}. 
Leveraging its efficiency in learning under temporally-changing data distributions through posterior updates, current efforts in the continual learning context explore approaches that integrate new information either assuming a continuous rate of change~\citep{nguyen2018variational, chang2022diagonal} or incorporating priors for changepoint detection~\citep{li2021detecting}.

\textbf{Probabilistic numerics.}
Probabilistic numerics~\citep{hennig2022probabilistic} is the study of numerical algorithms as Bayesian decision-makers. As numerical algorithms, such as optimization and linear algebra, are clearly central to deep learning, probabilistic numerics offers interesting prospects for making deep learning both more powerful and Bayesian. As one example, since deep training is now regularly I/O-bound for large models, active management of data loading, during training and UQ, is of increasing interest. Methods that quantify and control the information provided by individual computations, based on their effect on the BDL posterior, are showing promise as a formalism for algorithmic data processing in deep training~\citep{tatzel2023accelerating}, using probabilistic numerical linear algebra~\citep{NEURIPS2022_4683beb6} to select sparse informative `views' on the data.

\textbf{Singular learning theory.}
Singular learning theory~\citep[SLT;][]{wat09} investigates the relation between Bayesian losses, such as approximations of the marginal log-likelihood, and neural network loss functions, using principles from non-equilibrium statistical mechanics. Recent research has drawn connections between Bayesian methods and SLT~\citep{wei23}. 

\textbf{Conformal prediction.}
For UQ, alternatives such as conformal prediction have emerged as competitors to Bayesian methods and result in well-calibrated uncertainties~\citep{vovk05}. 
Deep learning models can be used to develop conformal prediction algorithms~\citep{meister2023} and, conversely,
conformal prediction methods can be used to quantify or calibrate uncertainty in deep learning models. 
A Bayesian approach to conformal prediction has started to emerge~\citep{hobbhahn2022fast,murphy2023probabilistic}, promising a synergistic approach that combines the strengths of Bayesian reasoning with the well-calibrated UQ offered by conformal prediction.

\textbf{LLMs as distributions.}
LLMs may be used flexibly as distribution objects in arbitrarily complex programs and workflows. By taking a Bayesian stance, several questions emerge for exploration. When multiple LLMs interact, how does one perform joint inference? What is an effective approach to marginalize over latent variables generated by LLMs, facilitating joint learning over such latent spaces?
Is it possible to adopt tools from computational statistics or approximate inference to perform various forms of reasoning with LLMs? And are there innovative ways to synergize small and large LLMs to amortize inferences just in time? 

\textbf{Meta-models.}
An intriguing prospect arises when contemplating whether BDL will parallel the trajectory of language models. Could one envision the development of a Bayesian meta-model within the BDL framework~\citep{krueger2017bayesian}? This meta-model, akin to language models, may be fine-tuned to multiple tasks, demonstrating competitive predictive performance across them, thus generalizing approaches in amortized inference~\citep{garnelo2018conditional, gordon2019convolutional,muller2021transformers}. 

\textbf{Sequential decision benchmarks.}
Standard image-based benchmarks focus exclusively on state-of-the-art predictive performance, where non-Bayesian deep learning algorithms typically have an advantage over BDL. To quantify predictive uncertainty, it is encouraged to shift attention to more thorough simulation studies or scientific applications focused on sequential learning and decision-making, such as experimental design, Bayesian optimization, active learning, or bandits. By prioritizing sequential problems in such contexts, researchers and practitioners can gain insights into how well a model generalizes to new and unseen data, how robust it is under uncertain conditions, and how effectively its uncertainty estimates can be utilized by decision makers in real-world scenarios.

\section{Final Remarks}\label{sec:final-remarks}

This paper has shown that modern deep learning faces a variety of persistent ethical, privacy, and safety issues, particularly when viewed in the context of different types of data, tasks, and performance metrics.
However, many of these issues can be overcome within the framework of Bayesian deep learning, building on foundational principles that have survived two and a half centuries of scientific and machine learning evolution.
While a number of technical challenges remain, there is a clear path forward that combines creativity and pragmatism to develop BDL approaches that match the data, hardware, and numerical advances of the twenty-first century, especially in the context of large-scale foundation models.
In a future where deep learning models seamlessly integrate into decision-making systems, BDL thus emerges as a crucial building block for more mature AI, adding an extra layer of reliability, safety, and trust.

%
%
%

\section*{Acknowledgements}

This work was supported in part by funding from the Eric and Wendy Schmidt Center at the Broad Institute of MIT and Harvard (MS).
JA is supported by ANR-21-JSTM-0001 grant.
VF was supported by a Branco Weiss Fellowship.
PH gratefully acknowledges financial support by the DFG Cluster of Excellence `Machine Learning - New Perspectives for Science', EXC 2064/1, project number 390727645; the German Federal Ministry of Education and Research (BMBF) through the T\"{u}bingen AI Center (FKZ: 01IS18039A); and funds from the Ministry of Science, Research and Arts of the State of Baden-W\"{u}rttemberg.
\care{JMHL acknowledges support from a Turing AI Fellowship under grant EP/V023756/1.}
SM acknowledges support from the National Science Foundation (NSF) under the NSF CAREER Award 2047418; NSF Grants 2003237 and 2007719, the Department of Energy, Office of Science under grant DE-SC0022331, as well as gifts from Disney and Qualcomm.
CN kindly acknowledges the support of EPSRC grants EP/V022636/1 and EP/Y028783/1.
\care{AGW is supported by NSF CAREER IIS-2145492, NSF I-DISRE 193471, NSF IIS-1910266, BigHat Biosciences and Capital One.}

The authors thank the ICML reviewers for their reviews and feedback.

%




\bibliographystyle{icml2024}

\newpage
\appendix
\onecolumn

\section{Background}
\label{sec:background}

This appendix provides background knowledge on several Bayesian methods that underpin Bayesian deep learning (BDL). It can be used as a self-contained introductory tutorial on the basics of Bayesian methodology and BDL. For a more detailed coverage, the reader is referred to the references provided herein.

\subsection{Laplace Approximations}

Laplace approximations constitute a method for constructing a Gaussian process (GP) posterior on the output of a neural network, leveraging automatic differentiation and numerical linear algebra. Consider a neural network $\fvec(\xvec,\thetavec)$ that maps input $\xvec$ and parameters $\thetavec\in\Rsp{\thetadim}$ (representing, for example, network weights and biases) to an output $\yvec$. The neural network is trained to find the parameters $\tilde{\thetavec}$ that minimize a regularized empirical risk function $\mathcal{L}(\thetavec)$ on supervised training data $\mathcal{D}=(\xvec_i,\yvec_i)_{i=1,\dots,\datadim}$.
\begin{equation*}
\tilde{\thetavec} = \operatorname*{argmin}_{\thetavec\in\Rsp{\thetadim}}\mathcal{L}(\thetavec) =
\sum_{i=1}^{\datadim} \ell(\yvec_i,\fvec(\xvec_i,\thetavec)) + r(\thetavec),
\end{equation*}
where $\ell$ and $r$ are a training loss and regularizer, respectively. The parameter value $\tilde{\thetavec}$ is found using the same approach as in non-Bayesian deep learning, employing stochastic optimization. It is possible to interpret the value $\tilde{\thetavec}$ obtained by training the neural network. In particular, minimizing $\mathcal{L}$ is equivalent to maximizing the exponential of negative $\mathcal{L}$, since the exponential function is strictly increasing:
\begin{align*}
    \tilde{\thetavec} &= \operatorname*{argmax}_{\thetavec\in\Rsp{\thetadim}}\exp(-\mathcal{L}(\thetavec))\\
    &= \operatorname*{argmax}_{\thetavec\in\Rsp{\thetadim}} \left( \prod_{i=1}^{\datadim} \exp(-\ell(\yvec_i,\fvec(\xvec_i,\thetavec))) \exp(-r(\thetavec))\right)\\
    &= \operatorname*{argmax}_{\thetavec\in\Rsp{\thetadim}} \prod_{i=1}^{\datadim} p(\yvec_i\mid \fvec(\xvec_i,\thetavec)) p(\thetavec) \\
    &= \operatorname*{argmax}_{\thetavec\in\Rsp{\thetadim}} p(\thetavec\mid \mathcal{D}),
\end{align*}
where $\ell$ is re-interpreted as a negative log-likelihood, and $r$ as a negative log-prior. This interpretation is valid for commonly used choices of these quantities in deep learning. The log-likelihood $\ell$ is commonly the logarithm of a distribution from the exponential family. Typical choices of $r$ are variants of the $l_2$ loss, such as the logarithm of a Gaussian prior on the parameters.

Under this interpretation, automatic differentiation can be used to compute a second-order Taylor approximation of $\mathcal{L}$ around $\tilde{\thetavec}$, and thus a Gaussian approximation for $p(\thetavec \mid \mathcal{D})$ can be acquired:
\begin{equation}
\label{eq:Laplace-weight-posterior}
\log p(\thetavec\mid \mathcal{D})\approx \mathcal{L}(\tilde{\thetavec}) + \frac{1}{2} (\thetavec-\tilde{\thetavec})^T \Psimat (\thetavec-\tilde{\thetavec}) = \log \mathcal{N}(\thetavec;\tilde{\thetavec}, -\Psimat^{-1}).
\end{equation}
$\Psimat$ is nominally the Hessian of $\mathcal{L}$. Due to its quadratic dependence on $\thetadim$, approximations are typically used. Of particular interest is the generalized Gauss-Newton (GGN) matrix $\Gmat$~\citep{schraudolph2007stochastic,martens2010deep},
\begin{equation}\label{eq:Psi}
\Psimat \approx \Gmat = \sum_{i=1}^{\datadim} \Jmat_{\tilde{\thetavec}}(\xvec_i) \left( \nabla_{\fvec} \nabla_{\fvec} ^T \ell(\yvec_i,\fvec(\xvec_i,\tilde{\thetavec})) \right) \Jmat_{\tilde{\thetavec}}^T (\xvec_i)  + \nabla\nabla^T r(\tilde{\thetavec}),
\end{equation}
which can be evaluated using the closed-form Hessian of the loss with respect to the logit inputs, and the Jacobian
\begin{equation*}
[\Jmat_{\tilde{\thetavec}}(\xvec_i)]_{a,b} =
\left. \frac{\partial f_b(\xvec_i ,\thetavec)}{ \partial \theta_{a}} \right|_{\thetavec = \tilde{\thetavec}}
\end{equation*}
of the neural network $\fvec$. This matrix has a low rank that allows efficient manipulation, such as computing the inverse required in~\Cref{eq:Psi}. To propagate this approximate belief on $\thetavec$ to the output of $\fvec$, it is common to linearize the network with respect to $\thetavec$ around $\tilde{\thetavec}$:
\begin{equation*}
\fvec(\xvec,\thetavec) \approx \fvec(\xvec,\tilde{\thetavec}) + (\thetavec - \tilde{\thetavec})^T \Jmat_{\tilde{\thetavec}}(\xvec).
\end{equation*}
Note that this approximation is made with respect to $\thetavec$; the neural network remains a non-linear function of its input $\xvec$.

Under this linearization, the posterior on $\fvec(\xvec)$ associated with the Gaussian posterior on $\thetavec$ of~\Cref{eq:Laplace-weight-posterior} is a GP:
\begin{equation*}
    p(\fvec(\xvec)\mid \mathcal{D}) \approx \mathcal{GP}\left(\fvec(\cdot),\fvec(\xvec,\tilde{\thetavec}), -\Jmat_{\tilde{\thetavec}}(\cdot) \Psimat^{-1} \Jmat_{\tilde{\thetavec}}(\cdot) \right).
\end{equation*}
The mean function of the GP corresponds to the trained neural network $\fvec(\cdot,\tilde{\thetavec})$ used in non-Bayesian deep learning. The GP kernel is the posterior version of the neural tangent kernel~\citep{jacot2018neural}. This concrete practical connection enables Laplace approximations to be used as a drop-in method in deep learning; the neural network is trained or a pre-trained one is used. Subsequently, the GGN matrix and Jacobian are computed. The trained neural network is then kept as a point estimate, now serving as the posterior mean of the GP, augmented with structured GP uncertainty. The computational overhead at training time is limited to the numerical linear algebra of cost that is linear in the training set size $\datadim$ and in the parameter space dimension $\thetadim$. At test time, inference for a given input $\xvec'$ requires one backward pass to compute the Jacobian $\Jmat_{\tilde{\thetavec}}(\xvec')$, resulting in a constant overhead compared to the forward pass needed to compute $\fvec(\xvec',\tilde{\thetavec})$.

An advantage of the Laplace approximation is that it enables to compute the marginal likelihood of the approximate posterior in closed form~\citep{immer2021scalable}, which can be used for Bayesian model selection in neural networks, for instance, for invariance learning~\citep{immer2022invariance}, linguistic probing of language models~\citep{immer2022probing}, or neural network pruning~\citep{dhahri2024shaving}. Thus, the Laplace approximation makes BDL more computationally feasible.

\subsection{Variational Inference}

Variational inference is an approach to approximate inference that seeks to avoid the intractability of exact inference by framing posterior inference as a variational optimization problem. Consider some stochastic parameters $\Thetamat$, data $\mathcal{D}$, a likelihood function $p(\mathcal{D} \mid \thetavec)$, a prior $p(\thetavec)$, and the posterior $p(\thetavec \mid \mathcal{D})$ given by
\begin{equation*}
    p(\thetavec \mid \mathcal{D})
    =
    \frac{p(\mathcal{D} \mid \thetavec) \, p(\thetavec)}{p(\mathcal{D})} .
\end{equation*}
Variational inference approximates $p(\thetavec \mid \mathcal{D})$ by solving the variational problem
\begin{equation}
\label{eq:variational_problem}
\min_{q_{\Thetamat}(\thetavec) \in \mathcal{Q}_{\Thetamat}} \DKL(q_{\Thetamat} \mid\mid p_{\thetavec |\mathcal{D}})
\end{equation}
with respect to a variational distribution $q(\thetavec)$ within some variational family of distributions $\mathcal{Q}_{\Thetamat}$~\citep{wainwright2008vi}. In expression~\eqref{eq:variational_problem}, $\DKL$ denotes the Kullback-Leibler (KL) divergence. Since the posterior $p(\thetavec \mid \mathcal{D})$ is the distribution to be approximated and as such is not accessible, the variational problem described by expression~\eqref{eq:variational_problem} cannot be solved directly. However, it can be shown that solving this variational problem is mathematically equivalent to maximizing the variational objective
\begin{equation*}
\mathcal{F}(q(\thetavec))
=
\mathbb{E}_{q(\thetavec)}[\log p(\mathcal{D} \mid \thetavec) ] - \DKL(q_{\Thetamat} \mid\mid p_{\Thetamat}) 
\end{equation*}
with respect to a variational distribution $q_{\Thetamat}(\thetavec) \in \mathcal{Q}_{\Thetamat}$.
Put another way, 
\begin{equation*}
    \min_{q_{\Thetamat}(\thetavec) \in \mathcal{Q}_{\Thetamat}} \DKL(q_{\Thetamat} \mid\mid p_{\thetavec \mid \mathcal{D}})
    \Longleftrightarrow
    \max_{q_{\Thetamat}(\thetavec) \in \mathcal{Q}_{\Thetamat}}  \mathcal{F}(q(\thetavec)) .
\end{equation*}
The variational objective $\mathcal{F}(q(\thetavec))$ is commonly referred to as the evidence lower bound (ELBO), since it can be shown that
\begin{equation*}
    \log p(\mathcal{D})
    =
    \mathcal{F}(q(\thetavec)) + \DKL(q_{\Thetamat} \mid\mid p_{\thetavec \mid\mathcal{D}}) ,
\end{equation*}
which, by non-negativity of the KL divergence, implies that $\log p(\mathcal{D}) \geq \mathcal{F}(q(\thetavec))$. So, the variational objective is a lower bound on the evidence, that is, on the log-marginal likelihood $\log p(\mathcal{D})$.
Finally, it is noted that $\log p(\mathcal{D}) = \mathcal{F}(q(\thetavec))$ if and only if $q(\thetavec) = p(\thetavec \mid \mathcal{D})$, which means that the ELBO is perfectly tight if and only if the variational distribution is equal to the posterior.

In general, variational inference is not guaranteed to converge to the posterior $p(\thetavec \mid \mathcal{D})$ unless the variational objective is convex in the variational parameters and the posterior is a member of the variational family, that is, $p(\thetavec \mid \mathcal{D}) \in \mathcal{Q}_{\Thetamat}$. Various approximate inference methods have been developed to solve the variational problem described by expression~\eqref{eq:variational_problem}. These methods make different assumptions about the variational family $\mathcal{Q}_{\Thetamat}$, and therefore result in different posterior approximations.

For variational inference with neural networks, two well-established methods are Monte Carlo dropout~\citep{gal2016dropout} and Gaussian mean-field variational inference~\citep[also referred to as Bayes-by-backprop;][]{blundell2015mfvi,graves2011practical}.
These methods are suited for stochastic mini-batch-based variational inference and can be scaled to large neural networks~\citep{hoffman13}.
Recent work on function-space variational inference~\citep[FSVI;][]{sun2018functional,rudner2022fsvi,rudner2022sfsvi} in Bayesian neural networks (BNNs) frames variational inference as optimization over induced functions, that is,
\begin{equation*}
    \min_{q_{\Fmat}(\fvec) \in \mathcal{Q}_{\Fmat}} \DKL(q_{\Fmat} \,||\, p_{\Fmat \mid \mathcal{D}})
\end{equation*}
for
\begin{equation*}
    p(\fvec \mid \mathcal{D})
    =
    \frac{p(\mathcal{D} \mid \fvec) \, p(\fvec)}{p(\mathcal{D})}
\end{equation*}
with a suitably defined prior distribution $p(\fvec)$ over functions. FSVI has been shown to result in state-of-the-art predictive uncertainty estimates in computer vision tasks~\citep{rudner2022fsvi}.

\subsection{Ensembles}

Deep ensembling refers to a procedure where a neural network architecture is re-trained multiple times with different initializations to find different parameter settings, and then the resulting predictive distributions at those parameter settings are averaged at test time~\citep{lakshminarayanan2017simple}. In practice, deep ensembles provide a simple approach to representing epistemic uncertainty by capturing the variability in model predictions.  This approach contrasts with more conventional Bayesian methods that involve sampling from the posterior distribution. Although unorthodox as an approximate inference procedure, deep ensembles often provide a closer approximation to the true posterior predictive distribution than many conventional approximate inference methods in deep learning~\citep{wilson2020bayesian, izmailov2021}, such as variational inference with a Gaussian approximate posterior.

In particular, one minimizes the standard loss for different initializations, which is often equivalent to minimizing a negative log-posterior $\log p(\thetavec\mid\mathcal{D})$ to obtain
\begin{equation*}
\tilde{\thetavec} =
\operatorname*{argmin}_{\thetavec\in\Rsp{\thetadim}}\mathcal{L}(\thetavec) =
\operatorname*{argmin}_{\thetavec\in\Rsp{\thetadim}}\left(-\log p(\thetavec\mid\mathcal{D})\right) =
\operatorname*{argmin}_{\thetavec\in\Rsp{\thetadim}}\left(-\log p(\mathcal{D}\mid\thetavec) - \log p(\thetavec)\right),
\end{equation*}
where $\thetavec\in\Rsp{\thetadim}$ are the neural network parameters, and $\mathcal{D}$ represents the training dataset. The negative log-likelihood $-\log p(\mathcal{D} \mid \thetavec)$ may correspond to cross-entropy loss, and a Gaussian prior $-\log p(\theta)$ corresponds to standard $\ell_2$ regularization or weight decay. After finding different local solutions $\tilde{\thetavec}_{1}, \ldots, \tilde{\thetavec}_{s}$ starting from different initializations, one averages the predictive distributions to make predictions given a test input $x'$:
\begin{align}
p(\yvec \mid \xvec',\mathcal{D}) = \frac{1}{s} \sum_{i=1}^{s} p(\yvec \mid \xvec', \tilde{\thetavec}_{i}) .
\label{eq: deepensemble}
\end{align}

Initially, the procedure of neural network ensembling at test time was not framed in probabilistic terms and was frequently described as a `non-Bayesian' alternative to standard approximate inference methods such as the Laplace approximation. However,~\Cref{eq: deepensemble} can be seen as approximating the true posterior predictive distribution
\begin{equation}
p(\yvec \mid \xvec',\mathcal{D}) =
\int p(\yvec \mid \xvec', \thetavec) \, p(\thetavec \mid \mathcal{D}) \, d\thetavec .
\label{eq: bma}
\end{equation}
There are different ways to interpret this predictive distribution. Several works have explored the connections between Bayesian inference and deep ensembles \citep{ciosek2020conservative,gustafsson2020evaluating,he2020bayesian,pearce2020uncertainty,wilson2020bayesian,izmailov2021,dangelo2021repulsive,dangelo2021stein}. One interpretation views~\Cref{eq: deepensemble} as a Monte Carlo approximation of~\Cref{eq: bma}, where the posterior of the parameters is represented as a set of point masses centered at different modes, which may be viewed as approximate posterior samples. However, this interpretation is not the most insightful. 

It is more enlightening to view approximate inference as the task of accurately approximating the integral in~\Cref{eq: bma}. From this perspective, the focus is not on collecting posterior samples. For a fixed computational budget, a Monte Carlo average of predictive distribution values based on exact posterior samples can provide a poor approximation of the integral relative to alternatives. A more compelling approach to numerical integration is to choose parameter values that represent typical points in the posterior, indicative of regions with significant posterior probability mass, and that yield diverse predictions on the test set. A heuristic to achieve this goal is to choose points corresponding to different posterior modes, as achieved by deep ensembles~\citep{wilson2020bayesian}. 
In practice, there is more functional variability across different posterior modes compared to samples in the vicinity of a single mode, such as the ones found from a variational Gaussian approximation of the posterior. 

These observations are corroborated in practice by experiments. Deep ensembles tend to provide a closer approximation to the posterior predictive distribution, represented by exhaustive Hamiltonian Monte Carlo (HMC) sampling, than conventional unimodal posterior approximations~\citep{izmailov2021}. The success of deep ensembles suggests that achieving a closer approximation to the posterior predictive distribution can lead to better predictive performance, highlighting the potential for further research. There are many natural ways to approximate the posterior predictive distribution. An obvious approach is to use a mixture of Gaussians centered at posterior modes, rather than a mixture of point masses. This approach has been found to approximate the posterior predictive distribution more closely and achieve better predictive performance in the NeurIPS 2021 approximate Bayesian inference competition~\citep{wilson2020bayesian, wilson2022evaluating, shen2024variational}.

A more general lesson to be extracted from these findings is that it is often not reasonable to consider whether a method is `Bayesian' as a binary; different approximate inference procedures fall onto a spectrum representing how closely they approximate the true posterior predictive distribution. Different methods provide better or worse approximations, depending on the model and the data. In the case where the parameter posterior is unimodal, deep ensembles are less useful as an inference procedure. On the other hand, if many modes are available and the modes correspond to functions that make different predictions, then deep ensembles are sensible as an approximate Bayesian inference procedure, especially under computational constraints when it is not feasible to represent many different parameter settings.\footnote{For more information on how deep ensembles facilitate approximate Bayesian inference, see the webpage \url{https://cims.nyu.edu/~andrewgw/deepensembles/}.}

\subsection{Posterior Sampling Algorithms}

Sampling algorithms, particularly Markov chain Monte Carlo (MCMC) methods, are widely used for Bayesian posterior inference. These algorithms work by constructing a Markov chain whose equilibrium distribution matches the desired (target) distribution. Updating the parameters by realizing a Markov chain yields samples from the target distribution, provided a sufficient number of updates are performed. Given a dataset $\mathcal{D}$, a model with parameters $\thetavec \in \Rsp{\thetadim}$, and a prior $p(\thetavec)$, the aim is to sample from the target posterior $p(\thetavec \mid \mathcal{D}) \propto \exp(-U(\thetavec))$, where the energy function is
\begin{equation*}
U(\thetavec) = -\sum_{\xvec \in \mathcal{D}} \log p(\xvec \mid \thetavec) - \log p(\thetavec).
\end{equation*}

Simulating the following continuous-time stochastic differential equation (SDE) produces samples with $p(\thetavec \mid \mathcal{D})$ as its stationary distribution:
\begin{equation}
\label{eq:langevin}
\mathrm{d}\thetavec = -\nabla U(\thetavec_t)\mathrm{d}t + 2\mathrm{d}B_t .
\end{equation}
$\nabla U(\thetavec)$ is the drift term of the SDE that guides the generated samples towards the posterior distribution, and $B_t$ is Brownian motion which introduces randomness into the process. The SDE in~\Cref{eq:langevin} is also known as the Langevin diffusion equation and is used as the basis of many Monte Carlo sampling algorithms~\citep{nemeth2021stochastic}. If the Langevin diffusion equation is considered over a small time interval $\alpha>0$, then a discrete-time version of it can be derived via the Euler-Maruyama approximation as
\begin{equation}
\label{eq:ula}
\thetavec_{k+1}=\thetavec_k-\alpha\nabla U(\thetavec_{k})+\sqrt{2\alpha}\boldsymbol{\xi}_{k+1},
\end{equation}
where \(\alpha>0\) is the step size parameter and $\boldsymbol{\xi}$ is standard Gaussian noise. This discrete-time algorithm is known as the unadjusted Langevin algorithm, or the Langevin Monte Carlo algorithm. However, unlike the continuous-time Langevin diffusion equation, the discrete-time unadjusted Langevin algorithm does not simulate samples with $p(\thetavec \mid \mathcal{D})$ as its stationary distribution, but instead produces samples that are only approximately drawn from $p(\thetavec \mid \mathcal{D})$. The discretization of the SDE leads to a bias in the posterior samples, which can be reduced by decreasing the step size parameter $\alpha$.

For large datasets, the unadjusted Langevin algorithm~\eqref{eq:ula} can be computationally expensive due to the need to sum over the entire dataset when evaluating $\nabla U(\thetavec)$. Stochastic gradient Langevin dynamics~\citep[SGLD;][]{welling2011bayesian} reduces the computational cost by using a stochastic gradient estimator $\nabla\tilde{U}$, an unbiased estimator of $\nabla U$ based on a subset of the dataset $\mathcal{D}$. SGLD has initiated a line of research on stochastic gradient MCMC (SG-MCMC) algorithms. It updates the vector of parameters $\thetavec$ at the $(k+1)$-th step according to
\begin{equation*}
\thetavec_{k+1} = \thetavec_{k} - \alpha \nabla \tilde{U}(\thetavec_{k}) + \sqrt{2\alpha} \boldsymbol{\xi}_{k+1}.
\end{equation*}
The key difference between SGLD and stochastic gradient descent (SGD) is the additional Gaussian noise in each step of SGLD, which allows it to characterize the full parameter posterior distribution rather than converging to a single point. 

Other notable variants of SG-MCMC include stochastic gradient HMC~\citep[SG-HMC;][]{chen2014stochastic}, which accelerates convergence using auxiliary momentum variables, and cyclical SG-MCMC~\citep{zhang2019cyclical}, which employs a cyclical step size schedule to efficiently explore multiple modes of the parameter posterior distribution. There have also been efforts to mitigate the bias in SG-MCMC methods by using Metropolis adjustments~\citep{zhang2020amagold, garriga2021exact}.

\subsection{Prior Specification}

The specification of prior $p(\thetavec \mid \mathcal{M})$ on a vector of parameters $\thetavec\in\Rsp{\thetadim}$ of a statistical model $\mathcal{M}$ has been a central part of Bayesian analysis, allowing to incorporate existing domain knowledge or expert opinion into statistical inference. A prior is called informative if it reflects such knowledge. Specific edge cases of informative priors include strongly informative priors, which dominate over the information coming from the observed data (likelihood), and weakly informative priors, which align with existing knowledge in a vague way so that the posterior is regularized to be data-informed and to be based on prior knowledge. In some cases, prior knowledge does not exist or a modeler does not want to rely on subjective knowledge. In such cases, uninformative or objective priors are used, where a common choice is a near-flat or even a uniform prior over the parameters. Another choice of objective prior worth mentioning is the reference prior, which is constructed to maximize some distance or divergence between the posterior and the chosen prior. Finally, in modern applications, priors are often selected to incorporate some desired properties into the model, such as regularization or sparsity.
The model $\mathcal{M}$ is considered herein to be a BNN. However, the priors discussed below are most commonly used in other statistical models, from which they have been typically adopted for BDL.

A common approach is to specify independent and identically distributed (i.i.d.) priors for the BNN parameters $\thetavec$. More specifically, a common default choice is to use a zero-centered isotropic Gaussian prior
\begin{equation*}
p(\thetavec \mid \mathcal{M}) = \prod_{i=1}^{\thetadim} \mathcal{N}(\theta_i;0,\sigma^2),
\end{equation*}
which corresponds to $l_2$ regularization in the sense of maximum a posteriori (MAP) solutions with $\lambda = 1/(2\sigma^2)$. Thus, the larger the prior variance, the less regularization is incorporated, and vice versa. Combined with specific activation functions, such as the logistic function, which is close to linear around 0, choosing a small $\sigma^2$ results in more linear behavior of the neurons and their compositions, while large $\sigma^2$ allows for more non-linear behavior. Thus, the popular approach of choosing standard Gaussian priors is not satisfactory in most cases and may lead to misspecified models. This, in turn, can cause the cold posterior effect that has been known to be the case for linear models~\citep{grunwald2017inconsistency}, but is also observed for BNNs~\citep{wenzel2020,fortuin2022bayesian,nabarro2022data}. For a specific problem, $\sigma^2$ can be chosen via hyper-parameter tuning or empirical Bayes. Moreover, a direct translation of the tuned $\sigma^2$ for some architecture (or, equivalently, $\lambda$ for frequentist neural networks) is possible.  Another approach is to impose an inverse-Gamma hyper-prior on $\sigma^2$, for example,  $\sigma^2\sim\Gamma^{-1}(\alpha,\beta)$, see~\citet{lampinen2001}. To incorporate prior dependencies between the parameters, i.i.d. Gaussian priors can be extended to multivariate normals with a zero mean vector and a covariance matrix $\Sigmamat$, i.e.,
\begin{equation*}
p(\thetavec \mid \mathcal{M}) = \mathcal{N}_{\thetadim}(\thetavec;\mathbf{0},\Sigmamat),
\end{equation*}
with the possibility to use an inverse-Wishart hyper-prior on $\Sigmamat$. Similarly to the i.i.d. Gaussian priors, independent Laplace priors can be used, i.e., 
\begin{equation*}
p(\thetavec \mid \mathcal{M}) = \prod_{i=1}^{\thetadim} \text{Laplace}(\theta_i;0,b), 
\end{equation*}
which in the MAP sense correspond to the $l_1$ regularization~\citep{williams1995bayesian} with $\lambda = b^{-1}$. Choosing the scale parameter $b$ can be done similarly to how $\sigma^2$ is chosen for Gaussian priors. Furthermore, Student-t priors have been used in the context of BNNs~\citep{neklyudov2018variance}. Heavy-tailed priors are possibly more robust towards model misspecification in the sense of the cold posterior effect~\citep{fortuin2022bayesian}. 

Another desirable property that is often integrated into BNNs is sparsity. Mixtures of Gaussians have been popular in this context, including a scale mixture of Gaussians prior~\citep{blundell2015},
\begin{equation*}
p(\thetavec \mid \mathcal{M}) =
\prod_{i=1}^{\thetadim} \left(\pi\mathcal{N}(\theta_i;0,\sigma_1^2) + (1-\pi)\mathcal{N}(\theta_i;0,\sigma_2^2)\right),
\end{equation*}
with $\sigma_1^2>\sigma_2^2$ and $\sigma_2^2\ll 1$.
Similarly, one can use horseshoe priors~\citep{carvalho2009handling}, 
\begin{equation*}
p(\thetavec \mid \mathcal{M}) =
\prod_{i=1}^{\thetadim} \mathcal{N}(\theta_i;0,\sigma^2\tau_i^2),
\end{equation*}
where $\tau_i$ is the local shrinkage parameter that has a half-Cauchy hyperprior $\tau_i \sim \text{C}^+(0,1)$, and $\sigma$ is the global shrinkage parameter.
Finally, another sparsity-inducing prior is the (improper) log-uniform prior~\citep{molchanov2017variational}, 
\begin{equation*}
p(\thetavec \mid \mathcal{M}) =
\prod_{i=1}^{\thetadim} \text{LogU}_\infty(\theta_i)\propto \prod_{i=1}^{\thetadim} \frac{1}{\theta_i},
\end{equation*}
and its proper counterpart~\citep{neklyudov2017structured}, 
\begin{equation*}
p(\thetavec \mid \mathcal{M})
\propto \prod_{i=1}^{\thetadim} \text{LogU}_\infty(\theta_i)\text{I}_{[a,b]}(\log \theta_i).
\end{equation*}

Some priors based on directional statistics have been explored for BNNs~\citep{sunde2023spherical}, but have not gained widespread adoption. Similarly, Jeffreys priors~\citep{ibrahim1991bayesian} have not been used extensively in this context. Although Zellner's g-priors~\citep{zellner1986assessing} and mixtures of g-priors~\citep{li2018mixtures} are highly popular in linear models, their application in BDL has only recently garnered attention~\citep{antoran2022linearised}.  

In Bayesian statistics, model uncertainty has been studied extensively for several decades~\citep{hoeting1998bayesian,hoeting1999bayesian,wasserman2000bayesian}. Within this framework, rather than having a single model $\mathcal{M}$, multiple BNN architectures $\{\mathcal{M}_1,\ldots,\mathcal{M}_t\}$ from a model space $\mathbb{M}$ are considered, making use of both $p(\thetavec \mid \mathcal{M})$ and $p(\mathcal{M})$. Recent research has focused on model uncertainty with respect to a model space  $\mathbb{M}$ defined by different patterns of weight inclusion~\citep{hubin2019combining,skaaret2023sparsifying}, resulting in $2^{\thetadim}$ models in $\mathbb{M}$. This requires additional model priors. If a model $\mathcal{M} = (\gamma_1,\ldots,\gamma_{\thetadim})$ with $\gamma_i\in\{0,1\}, i \in \{1,\ldots,\thetadim\}$, is assumed, then~\citet{hubin2019combining} and~\citet{skaaret2023sparsifying} propose
\begin{equation*}
p(\mathcal{M}) = \prod_{i=1}^{\thetadim}\text{Bernoulli}(\gamma_i;\rho_i),
\end{equation*}
with $\rho_i$ being the prior inclusion probability for a specific weight. Similarly,~\citet{hubin2024sparse} propose to use
\begin{equation*}
p(\mathcal{M}) \propto \prod_{i=1}^{\thetadim}\text{BetaBinomial}(\gamma_i;1,a_i,b_i).
\end{equation*}
These two types of prior are common in the Bayesian model-averaging literature~\citep{hoeting1999bayesian, corani2015credal}. However, more advanced model priors that incorporate dependencies between parameter inclusions through, for example, Dirichlet process hyper-priors~\citep{grun2021identifying} or dilution priors~\citep{george2010dilution}, have not yet been studied in the context of BDL. It is also noteworthy that, for model priors, inclusion probabilities for specific covariates for the input layer can be adjusted by experts according to prior knowledge, thus allowing the incorporation of domain-specific information into inference for BNNs. 

Incorporating prior knowledge directly into parameter priors presents a challenge in general. 
However, recent advances in probabilistic modeling for neural networks have shown that incorporating prior knowledge is possible. One approach involves leveraging auxiliary objectives to create data-driven priors~\citep{lechuga2023m2d2,rudner2023fseb,rudner2024uap,sam2024}. Another approach to specifying meaningful priors for neural networks is to adopt a function-space perspective. In this approach, BNNs generate a distribution $p(\fvec)$ over functions when sampling from parameter priors. A functional prior, such as a Gaussian process $p(\fvec) = \mathcal{GP}(\muvec (\cdot),\Kmat (\cdot,\cdot))$, can then be assumed for the function-space output, allowing the incorporation of expert knowledge about the mean and covariance functions for a specific phenomenon of interest. However, a direct application of this type of functional prior can be problematic due to potential mismatches between the support of the GP and the outputs of the BNN; see~\citet{burt2020understanding,rudner2022fsvi}. For the same reason, using the KL divergence to pre-train the priors over the parameters to match the chosen GP prior is problematic.~\citet{rudner2022fsvi} resolve this issue by considering a KL divergence between distributions over functions that are absolutely continuous to one another by design.~\citet{tran2022} also tackle this challenge by using the 1-Wasserstein distance instead of the KL divergence to learn the parameters of the priors in the weight space that match a chosen GP prior.

This section offers only a concise glimpse into the extensive literature on priors for BNNs. For a more comprehensive understanding and a relatively recent review, the reader is referred to~\citet{fortuin2022priors}. 

\subsection{Deep Kernel Processes}

A deep kernel process~\citep[DKP;][]{aitchison2021deep} is a Bayesian model that places a prior on a deep sequence of kernel representations. This is a change of perspective compared to other deep Bayesian models such as deep GPs~\citep[DGPs;][]{damianou2013} or BNNs~\citep{mackay1995bayesian}, which place priors over intermediate layer features or weights. DKPs are equivalent to DGPs whenever the kernel function is isotropic (such as the radial basis function and Mat\'{e}rn kernel). To illustrate this, consider a DGP with an isotropic kernel $\Cmat$, where each layer is modeled as a multivariate Gaussian conditioned on the preceding layer,
\begin{subequations}
\label{eq:dgp_defn}
\begin{align}
  \Fmat_0 &= \mathbf{X},\\
  g(\Fmat_\layeridx\mid \Fmat_{\layeridx-1}) &= \prod_{\featidx=1}^{\numfeats{\layeridx}} \mathcal{N}(\fvec_{\featidx,\layeridx}; \mathbf{0}, \Cmat(\Fmat_{\layeridx})),\\
  g(\Ymat\mid \Fmat_{\numhlayers+1}) &= \prod_{\featidx=1}^{\numfeats{\layeridx}} \mathcal{N}(\yvec_{\featidx}; \fvec_{\featidx,\numhlayers+1}, \sigma^2 \mathbf{I}).
\end{align}
\end{subequations}
Here, $\Fmat_{\layeridx}\in\Rsp{\datadim\times \numfeats{\layeridx}}$ are the feature representations in each intermediate layer $\layeridx\in\{1,\ldots,\numhlayers\}$, $\Xmat\in\Rsp{\datadim\times \numfeats{0}}$ are the inputs, $\Ymat\in\Rsp{\datadim\times \numnfeats{\numhlayers+1}}$ are the labels, and $\datadim$ are the number of data points. The subscript $\featidx$ denotes individual features, so that $\fvec_{\featidx,\layeridx}\in\Rsp{\datadim}$ is the $\featidx$-th feature at layer $\layeridx$, and $\yvec_{\featidx}\in\Rsp{\datadim}$ is the $\featidx$-th output for all data points. $\numfeats{\layeridx}$ is the number of features per data point at layer $\layeridx$, or `the width of layer' $\layeridx$. To obtain a kernel process, one needs to consider covariance matrices. So, for each layer, the Gram matrix $\Gmat_{\layeridx} = \Fmat_{\layeridx}\Fmat_{\layeridx}^T/ \numfeats{\layeridx}\in\Rsp{\datadim\times\datadim}$ is defined. Since $\Gmat_{\layeridx}$ is the outer product of i.i.d. Gaussian samples with covariance $\Cmat (\Fmat_{\layeridx})$, it must be Wishart distributed,
\begin{equation*}
g(\Gmat_{\layeridx} \mid \Fmat_{\layeridx-1}) =
\mathcal{W}(\Gmat_{\layeridx}; \Cmat (\Fmat_{\layeridx-1})/\numfeats{\layeridx}, \numfeats{\layeridx}).
\end{equation*}
Furthermore, by the isotropic assumption, there is a function $\Kmat (\cdot)$ over Gram matrices such that $\Kmat(\Gmat_{\layeridx}) = \Cmat(\Fmat_{\layeridx})$; see~\citet{aitchison2021deep}. 
The ability to define the kernel function in terms of Gram matrices means that it is possible to write the DGP in~\Cref{eq:dgp_defn} as a DKP with Wishart priors,
\begin{subequations}\label{eq:dkp_dgp}
    \begin{align}
g(\Gmat_1\mid\Xmat) &=
\mathcal{W}(\Gmat_1; \Xmat\Xmat^T/\numfeats{0}, \numfeats{0}),\\
g(\Gmat_{\layeridx}\mid\Gmat_{\layeridx-1}) &=
\mathcal{W}(\Gmat_{\layeridx}; \Kmat(\Gmat_{\layeridx-1})/\numfeats{\layeridx}, \numfeats{\layeridx}),\\
g(\yvec_{\featidx}\mid\Gmat_{\numhlayers}) &=
\mathcal{N}(\yvec_{\featidx}; \mathbf{0}, \Kmat(\Gmat_{\numhlayers}) + \sigma^2\mathbf{I}).
    \end{align}
\end{subequations}
Since~\Cref{eq:dkp_dgp} places Wishart priors on the intermediate Gram matrix representations, the resulting process is known as a deep Wishart process~\citep[DWP;][]{aitchison2021deep}. Deep inverse Wishart processes~\citep[DIWPs;][]{aitchison2021deep} are defined using inverse Wishart process priors over kernels~\citep{shah2014student} instead.

Similarly to other deep Bayesian models, closed-form inference of general DKPs is not possible.~\citet{aitchison2021deep},~\citet{ober2021variational} and~\citet{ober2023improved} have developed approximate posteriors over Gram matrices for DWPs and DIWPs to allow for variational inference. However, despite the use of approximate posteriors, the computational cost of training a DWP or DIWP remains considerable. This is because the number of parameters scales quadratically with the number $\datadim$ of data points, and evaluating the log-probabilities of their approximate posteriors (necessary when evaluating the ELBO) scales cubically with $\datadim$. To address this scalability challenge, inducing point approximations offer a solution. In particular, global inducing point methods~\citep{ober2021global} enable the training of DKPs with linear scaling in the number of data points.

Using inducing point schemes,~\citet{ober2023improved} have empirically demonstrated that approximate posteriors for DWPs perform better than DGP approximate posteriors.~\citet{aitchison2021deep} argue that DWPs are expected to perform better due to BNN and DGP priors and posteriors being highly multimodal. In particular, rotation and permutation symmetries in features or weights are not adequately accounted for by common BNN and DGP approximate posteriors. DKPs sidestep this multimodality issue, as Gram matrices inherently avoid these symmetries; an arbitrary rotation or permutation in feature space can be represented by the mapping $\Fmat\mapsto \mathbf{FU}$,  where $\mathbf{U}$ is unitary, yet the corresponding Gram matrix is invariant under this transformation since $\Gmat = \mathbf{FF}^T \mapsto(\mathbf{FU})(\mathbf{FU})^T = \Gmat$.

\subsection{Deep Kernel Machines}

Deep kernel machines~\citep[DKMs;][]{yang2023theory} are an infinite-width analog of DKPs. They have practical benefits in being easier to implement and cheaper to train, and also theoretical benefits as they can be linked to the existing infinite-width neural network literature. However, DKMs are not strictly Bayesian.
Usually, taking an infinite-width limit of a DKP or DGP results in a neural network Gaussian process~\citep[NNGP;][]{lee2017deep,agrawal2020}. The infinite-width limit is taken carefully, in such a way so as to retain flexibility in intermediate Gram representations.

A DKM can be obtained from the DGP of~\Cref{eq:dgp_defn} as follows.
Consider the following approximate posterior for the features in each intermediate layer $\layeridx\in\{1,\ldots,\numhlayers\}$:
\begin{equation*}
h(\Fmat_{\layeridx}) = \prod_{\featidx=1}^{\numfeats{\layeridx}} \mathcal{N}(\mathbf{f}_{\featidx,\layeridx}; \mathbf{0}, \Gmat_{\layeridx}) .
\end{equation*}
Moreover, consider a standard GP approximate posterior for the final layer:
\begin{equation*}
h(\Fmat_{\numhlayers+1}) = \prod_{\featidx=1}^{\numfeats{\numhlayers+1}} \mathcal{N}(\fvec_{\featidx,\numhlayers+1}; \muvec_{\featidx}, \Sigmamat).
\end{equation*}
Here, $\Gmat_1, \ldots,\Gmat_{\numhlayers},\muvec_1,\ldots,\muvec_{\numfeats{\numhlayers+1}}$, and $\Sigmamat$ are variational parameters.
Although this approximate posterior family may seem restrictive, the intermediate layer part contains the true posterior in the infinite-width limit; see Appendix E of~\citet{yang2023theory}. A lower bound for the marginal likelihood can be obtained via the ELBO
\begin{equation*}
    \text{ELBO} =
    \sum_{\featidx=1}^{\numfeats{\numhlayers+1}} \big\{
    \mathbb{E}_{h(\Fmat_{\numhlayers+1})}\big[\log g(\yvec_{\featidx} \mid \fvec_{\featidx,\numhlayers+1})\big] - \DKL(h(\fvec_{\featidx,\numhlayers+1})\mid\mid g(\fvec_{\featidx,\numhlayers+1}\mid \Fmat_{\numhlayers}))
    \big\} \nonumber
    - \sum_{\layeridx=1}^{\numhlayers}\beta_{\layeridx} \numfeats{\layeridx} \DKL(
    h(\fvec_{\layeridx})\mid\mid
    g(\mathbf{f}_{\layeridx}\mid \Fmat_{\layeridx-1})),
\end{equation*}
where tempering is employed using the parameter $\beta_{\layeridx}$.
As with DWPs, an isotropic kernel function is assumed, which means that $\Cmat(\mathbf{F}_{\layeridx}) = \Kmat (\mathbf{G}_{\layeridx})$. As the intermediate layers become wider by sending $\featc\rightarrow\infty$ with $\numfeats{\layeridx} = \featc \numnfeats{\layeridx}$, the dependency on $\Fmat_{\layeridx}$ disappears. If no tempering is applied (that is, $\beta_{\layeridx} = 1$), then the following objective is recovered:
\begin{equation}
\label{eq:nngp_elbo}
\frac{\text{ELBO}}{\featc}\rightarrow - \sum_{\layeridx=1}^{\numhlayers}\numnfeats{\layeridx}
\DKL (\mathcal{N}(\mathbf{0}, \Gmat_{\layeridx})\mid\mid \mathcal{N}(\mathbf{0}, \Kmat (\Gmat_{\layeridx-1})).
\end{equation}
The objective~\eqref{eq:nngp_elbo} is maximized when $\Gmat_{\layeridx} = \Kmat (\Gmat_{\layeridx-1})$, which is the same as the corresponding NNGP~\citep{lee2017deep,agrawal2020}. If tempering is carried out according to the width with $\beta_{\layeridx} = 1/\featc$, then the following objective is obtained:
\begin{equation}
\label{eq:dkm_obj}
\begin{aligned}
\text{ELBO} \rightarrow
&   \sum_{\featidx=1}^{\numfeats{\numhlayers+1}}
\mathbb{E}_{h(\Fmat_{\numhlayers+1})}\big[\log g(\yvec_{\featidx} \mid \fvec_{\featidx,\numhlayers+1})\big]\\ 
&-\sum_{\featidx=1}^{\numfeats{\numhlayers+1}}
\DKL(\mathcal{N}(\muvec_{\featidx}, \Sigmamat)\mid\mid \mathcal{N}(\mathbf{0}, \Kmat (\Gmat_{\numhlayers}))\\
&- \sum_{\layeridx=1}^{\numhlayers}\numnfeats{\layeridx}
\DKL(\mathcal{N}(\mathbf{0}, \Gmat_{\layeridx})\mid\mid
\mathcal{N}(\mathbf{0}, \Kmat (\Gmat_{\layeridx-1})) + \mathrm{constant}.
\end{aligned}
\end{equation}
A model that optimizes the objective~\eqref{eq:dkm_obj} is called a DKM, and~\eqref{eq:dkm_obj} is known as the DKM objective. In the limit, the DKM objective does not depend on intermediate features $\Fmat_{\layeridx}$, which means that learned representations in a DKM are described entirely by deterministic Gram matrices $\Gmat_1,\ldots,\Gmat_{\numhlayers}$.
To interpret the DKM objective, notice that the likelihood term encourages data fitting, and the KL divergences regularize the model toward the NNGP~\citep{lee2017deep,agrawal2020}. The amount of representation learning in the DKM can be controlled by varying the $\numnfeats{\layeridx}$ parameters. In contrast, the lack of likelihood term in the NNGP objective~\eqref{eq:nngp_elbo} prevents representation learning from occurring in the NNGP; intermediate Gram matrices are fixed and depend only on the input data.

Similarly to DKP objectives, the DKM objective is computationally infeasible to optimize for large datasets, with cubic scaling in the number of data points. However,~\citet{yang2023theory} have shown that the DKM objective can be optimized with linear scaling if global inducing point methods are used.
DKMs have been extended to convolutional architectures, achieving performance nearly on par with neural networks on CIFAR-10~\citep{milsom2023convolutional}.

\section{Diagnostics, Metrics and Benchmarks}\label{sec:diagnostics}

Currently, there is a lack of convergence and performance metrics specifically for the needs of BDL. Developing such tools can help identify the goals in BDL as well as assess their progress. 
Besides, the choice of evaluation metrics, datasets and benchmarks lack consensus in the BDL community which reflects a difficulty in clearly defining the goals of BDL in a field traditionally viewed through frequentist lens, specifically in terms of performance on test data. 
Many of the general Bayesian diagnostic and evaluation approaches are proposed through Bayesian workflow \citep{gelman2020bayesian}. This appendix discusses the most relevant approaches for BDL. 

\textbf{Convergence diagnostics in parameter space.}
The analysis of convergence and sampling efficiency~\citep{gelman2013, vehtari2021} for SG-MCMC sampling becomes a delicate matter, which is currently bypassed by a rather simplistic analysis of these quantities using summary statistics of predictive distributions. 
More generally, verifying the convergence of inference algorithms in the high-dimensional and multimodal settings of BDL models is not straightforward.
Convergence checks designed for BNNs need to be further studied.

\textbf{Performance metrics in predictive space.}
BDL and GP literature often focus on the mean of the predictive distribution, overlooking the analysis of variance of the predictive distribution. Some performance metrics are commonly used to assess variance levels, for example, by evaluating the log-likelihood or the entropy of predictions for test data~\citep{rudner2022fsvi,rudner2023fseb}. However, a systematic way to characterize the predictive uncertainty in BDL inference (apart from binary classification problems where AUROC and AUPRC are widely used) is currently lacking~\citep{arbel2023}. The challenge of setting metrics for the assessment of epistemic and aleatoric uncertainty slows the progress in BDL and could potentially be addressed by establishing widely accepted benchmarks for BDL methods. 

\textbf{Performance metrics in misspecified settings.}
Addressing challenges related to distribution shift and test data performance requires the development of robust performance metrics. To establish BDL model reliability under distribution shift, tighter generalization bounds, such as those provided by the PAC-Bayes framework~\citep{langford2002pac, parrado2012pac}, are crucial to obtain probabilistic guarantees on model performance. Furthermore, in misspecified settings, evaluating calibration becomes paramount. Innovative techniques, such as two-stage calibration~\citep{guo2017calibration} and conformal prediction~\citep{papadopoulos2007conformal} or its Bayesian counterpart~\citep{hobbhahn2022fast}, offer practical solutions by refining predicted probabilities and quantifying predictive uncertainty, respectively. These approaches collectively contribute to a more comprehensive evaluation of model performance in scenarios where the underlying assumptions may not align with the true data distribution.

\textbf{Probabilistic treatment of datasets.}
Probabilistic treatments of data as a first-class citizen that can be reasoned about in BNNs seem promising. Such probabilistic approaches may help create more focused and useful datasets to represent the knowledge contained in vast data sources, improving the ability to train and maintain large models. 

\section{Software Usability}
\label{sec:software}

Applying a BDL approach to a real-world problem is still a more complex endeavor than opting for an off-the-shelf standard deep learning solution, which limits the real-world adoption of BDL. 
Software development is key to encouraging deep learning practitioners to use Bayesian methods. 
More generally, there is a need for software that would make it easier for practitioners to try BDL in their projects. The use of BDL must become competitive in human effort with standard deep learning. 

Some efforts have been made to develop software packages, libraries or probabilistic programming languages (PPLs) on top of deep learning frameworks. \texttt{bayesianize}~\citep{ritter2021}, \texttt{bnn\_priors}~\citep{fortuin2021bnnpriors}, \texttt{Laplace}~\citep{daxberger2021b}, \texttt{Pyro}~\citep{bingham2019} and \texttt{TyXe}~\citep{ritter2022} are software species built on \texttt{PyTorch}, \texttt{TensorFlow Probability} is a library built on \texttt{TensorFlow}, and \texttt{Fortuna}~\citep{detommaso2023fortuna} is a library built on JAX. It would help to make further progress with contributions from the probabilistic programming community. 

PPLs, such as \texttt{Pyro}, play a role in simplifying the application of probabilistic reasoning to deep learning. In fact, abstractions of the probabilistic treatment of NNs in a PPL, such as those performed in the BDL library \texttt{TyXe}, can simplify the application of priors and inference techniques to arbitrary NNs, as demonstrated in a variety of models implemented in \texttt{TyXe}. Porting such ideas to modern problem settings involving LLMs and more bespoke probabilistic structures would enable the use of BDL in real-world problems. 

Contemporary deep learning pushes the limits of scale in all dimensions: datasets, parameter spaces, and structured function-valued output. For point estimation, the community has been developing array-centric programming paradigms that allow sharding, partial evaluations, currying, and more. BDL should be able to map these ideas to develop analogous software. 

\section{Topical Developments}
\label{sec:topical}

This appendix provides topical or specialized areas of BDL for future development. These include BDL for human-AI interaction, lifelong and decentralized learning, Bayesian reinforcement learning (RL), and domain-specific BDL models.

\textbf{Human-AI interaction and explainable AI.}
Enabling AI systems to communicate and explain their uncertainty can build trust and improve the interaction between AI systems and humans. While efforts by the community have been made to explain the predictions of DNNs, recent efforts aim to explain the uncertainty of BDL methods~\citep{antoran2021getting, bhatt2021}. Understanding which input patterns are responsible for high predictive uncertainty can build trust in AI systems and can provide insights about input regions where data is sparse. For example, when training a loan default predictor, a data scientist can identify population subgroups (by age, gender, or race) underrepresented in the training data. Collecting more data from these groups can lead to more accurate predictions for a wider range of clients. 

\textbf{Lifelong and decentralized learning.}
A contemporary research direction is to go beyond the `static' train-test framework and focus on `dynamic' problems where the test set is not known. This includes cases where predictive performance, robustness and safety are important and there are realistic constraints on the infrastructure. Two such problems are lifelong learning and decentralized learning. Focusing on such problems is expected to lead to a new regime in which Bayesian ideas can be useful for deep learning. 

\textbf{Efficient exploration in RL.}
RL is an area where BDL has shown potential. As an example, Thompson sampling (TS) is known to be a commonly used heuristic for decision making that `randomly selects an action, according to the probability that it is optimal'~\citep{russo2018tutorial}. TS balances exploration with exploitation and in its exact form requires sampling from the Bayesian posterior. In practice, approximations are often used, and recent work has shown that the quality of the resulting multivariate joint predictive distribution over multiple test inputs is important for decision-making~\citep{wen2021predictions, osband2023}. This is relevant, as typical Bayesian and non-Bayesian methods are commonly evaluated by assessing the quality of marginal predictions over individual test inputs, ignoring potential dependencies~\citep{osband2022neural}. While deep ensembles are a typical baseline for capturing uncertainty, BDL methods based on the last-layer Laplace approximation can outperform deep ensembles in the quality of joint multivariate predictions~\citep{antoran2023}. Developing methods that achieve trade-offs between computational cost and the quality of their joint multivariate predictions is an area where further research is needed~\citep{osband2023}. Another active area of research at the intersection of RL and BDL aims to produce accurate posterior approximations of value functions (for example, Q functions) given data from interactions with an environment~\citep{janz2019successor}. This setting is different from typical Bayesian supervised learning as, in this case, the output of value functions is not directly observed, and only rewards are available. 

\textbf{\care{Computer vision.}}
\care{BDL approaches to computer vision tasks have been developed. For instance,~\citet{kou2024} employ BDL in diffusion models to construct a pixel-wise uncertainty estimator for image generation.~\citet{goli2023} use BDL to evaluate uncertainty in pre-trained neural radiance fields in the context of computer graphics. Future research in BDL for computer vision may focus on improving predictive performance and further developing UQ methods. Computer vision, along with natural language processing, constitute applications that may promote the adoption of BDL.}

\textbf{Domain-specific BDL models.}
There are many opportunities to develop Bayesian methods in combination with deep learning models that are tailored for specific domains, taking into account the characteristics of the data and the tasks involved. This can involve exploring hierarchical models, transfer learning, or meta-learning approaches. An example is molecular property prediction, where many different datasets are available, but each of them has limited available data~\citep{klarner2023qsavi}. There is scope to combine deep learning models that learn molecular feature representations with Bayesian methods that receive those representations as inputs. The latter methods can capture uncertainty and make predictions in data-limited settings for each individual task, while the deep learning features are shared across tasks. 






\end{document}